\documentclass[runningheads, envcountsame, a4paper]{llncs}
\usepackage[pdftex]{graphicx}

\usepackage{multirow}
\usepackage{epsfig}
\usepackage{amsmath}
\usepackage{amssymb}
\usepackage{float}
\usepackage{tasks}
\usepackage{comment}
\usepackage[utf8]{inputenc}
\usepackage[super]{nth}

\usepackage[misc]{ifsym}
\usepackage[hyphens]{url}
\usepackage{pgfplots} 

\usepackage{mathtools}
\usepackage{subcaption}
\newcommand{\etal}{~et~al.~}
\newcommand{\quotes}[1]{``#1''}

\def \plotHeight {0.4\linewidth}

\def \triplePlotWidth {\linewidth}

\def \triplePlotAxesFontSize {\small}
\def \triplePlotTicksFontSize {\small}

\begin{document}
\toctitle{Attentive Multi-Task Deep Reinforcement Learning}
\title{Attentive Multi-Task Deep Reinforcement Learning}
%
%
\tocauthor{Timo~Br\"am, Gino~Brunner, Oliver~Richter, Roger~Wattenhofer}
\author{Timo Br\"am\and Gino Brunner {\Letter} \orcidID{0000-0002-4341-2940} \and \\Oliver Richter {\Letter} \orcidID{0000-0001-7886-5176} \and Roger Wattenhofer\thanks{Authors listed in alphabetical order.}}
%
%
\authorrunning{T. Br\"am et al.}
%
\institute{Department of Information Technology and Electrical Engineering \\ ETH Zurich \\ Switzerland \\
\email{$\{$brunnegi,richtero,wattenhofer$\}$@ethz.ch \\ braem.timo@gmail.com}}
\maketitle              
\setcounter{footnote}{0}
\begin{abstract}
Sharing knowledge between tasks is vital for efficient learning in a multi-task setting.
However, most research so far has focused on the easier case where knowledge transfer is not harmful, i.e., where knowledge from one task cannot negatively impact the performance on another task.
In contrast, we present an approach to multi-task deep reinforcement learning based on attention that does not require any a-priori assumptions about the relationships between tasks. Our attention network automatically groups task knowledge into sub-networks on a state level granularity. It thereby achieves positive knowledge transfer if possible, and avoids negative transfer in cases where tasks interfere. We test our algorithm against two state-of-the-art multi-task/transfer learning approaches and show comparable or superior performance while requiring fewer network parameters.
\end{abstract}

\section{Introduction}
Humans are often excellent role models for machines. Unlike machines, humans have been interacting with their environment since time immemorial, and this extensive experience should not be ignored. So how are we humans learning, and what can machines learn from us?

First, humans learn with a limited amount of training data, as we cannot afford to first train for an unreasonably long time before becoming active. 
Also, we usually do not require labeled training data, but instead rely on experience gained from interactions with our world.
This situation is well represented by the reinforcement learning paradigm: We observe the environment and take actions to hopefully maximize our cumulative reward.
Second, humans learn many tasks concurrently, not only because there is no time to learn all possible tasks sequentially, but also because tasks are often similar in nature, and useful strategies can be transferred between comparable tasks. This is a fundamental aspect of intelligence known as multi-task learning. 
Third, our brain would be overwhelmed if it had to focus on all skills acquired over the span of our lives at every point in time. Therefore, we focus our attention to a set of skills useful at the moment. If we were not able to relate similar tasks and attend to skills based on extrinsic or intrinsic cues, our brain would not be able to learn much.
Recent advances in neuroscience \cite{human_attention} also suggest that the attention mechanisms of humans are themselves learned through reinforcement learning.

In this paper, we investigate the combination of these three paradigms, in other words, we study \emph{attentive multi-task deep reinforcement learning}. 
More specifically, we employ the insight of human attention by developing a simple yet effective architecture for model-free multi-task reinforcement learning. We use a neural network based attention mechanism to focus on sub-networks depending on the current state of the environment and the task to be solved. Most recent work \cite{Distral,DARLA,shared_dynamics,sf_4} in the multi-task/transfer deep reinforcement learning setting capitalize on some shared property between tasks. In contrast, our approach makes no assumptions about the similarity between tasks.
Instead, possible relations are automatically inferred during training.

An additional advantage of using an attention based architecture is that unrelated tasks can effectively be separated and learned in different sub-parts of the architecture. We thereby automatically embrace the \emph{negative transfer} problem (the effect that training one task might actually harm performance on another task) which most related approaches omit in their evaluation.
We show that our approach scales economically with an increasing number of tasks as the attention mechanism automatically learns to group related skills in the same part of the architecture.
We back our claims by comparing against two state of the art algorithms \cite{Distral,PNN} on a large set of grid world tasks with different amounts of transferable knowledge. We show that our method scales better in the number of parameters per task, while achieving comparable or superior performance in terms of steps to convergence. Especially, when the action spaces of the tasks are not aligned we outperform \cite{Distral,PNN}.\footnote{To stimulate future research in this area, our source code is available at: \url{https://github.com/braemt/attentive-multi-task-deep-reinforcement-learning}.}

\section{Related Work}

Transfer learning in classical reinforcement learning \cite{DBLP:journals/jmlr/TaylorS09} is a well established research area.
Even though Lin~\cite{early_NN_RL} already used neural networks in combination with reinforcement learning,  a renewed interest in this combination came with the recent success on Atari~(DQN, \cite{DQN}), followed by
an increased interest in developing transfer learning techniques specific to deep learning.
Parisotto\etal\cite{actor-mimic} train a neural network to predict the features and outputs of several expert DQNs and use multi-task network weights as initialization for a target task DQN. Rusu\etal\cite{DBLP:journals/corr/RusuCGDKPMKH15} use a single network to match expert DQN policies from different games by policy distillation. Yin\etal\cite{DBLP:conf/aaai/YinP17} improve policy distillation by making the convolutional layers task specific and by using hierarchical experience replay. Schmitt\etal\cite{KickstartingDRL} also build on the idea of policy distillation but additionally propose to anneal the teacher signal such that the student can surpass the teacher's performance. Further, in \cite{DQfD,DBLP:journals/corr/abs-1802-05313,Ape-X_DQfD} knowledge is transferred from human expert demonstrations, while the algorithm of Aytar\etal\cite{DBLP:journals/corr/abs-1805-11592} learns from YouTube video demonstrations. Gupta\etal\cite{DBLP:journals/corr/GuptaDLAL17} transfer knowledge from source to target agent by training matched feature spaces. Closely related to our approach is the work of Rajendran\etal\cite{A2T} who also incorporate several sub-networks and an attention mechanism to transfer knowledge from an expert network. In contrast to the architecture described in \cite{A2T} and all related work mentioned so far, our algorithm learns multiple tasks simultaneously from scratch, without guidance from any demonstrations or experts. This makes our approach self-sustained and as such more general than mentioned related work.

Glatt\etal\cite{DBLP:conf/bracis/GlattSC16} train a DQN on a source task and investigate how the learned weights, which are used as initialization for a target task, alter the performance. In a similar manner, \cite{multi-task-A3C,IMPALA,PopArt-IMPALA} show that some transfer is possible by simply training one network on multiple tasks. However, since these algorithms do not incorporate any task-specific weights, the best that can be done is to interpolate between conflicting tasks. In contrast, our method allows conflicting tasks to be learned in separate networks.

One interesting line of research \cite{shared_dynamics,sf_1,sf_2,sf_4,sf_8} capitalizes on transferring knowledge based on successor features, i.e., shared environment dynamics. In contrast, our method does not rely on shared environment dynamics nor action alignment across tasks.

Czarnecki\etal\cite{czarnecki2018mix} use multiple networks similar to our approach. However, their focus is on automated curriculum learning. Therefore they adjust the policy mixing weights through population based training~\cite{populationbasedtraining} while we learn attention weights conditioned on the task state.

Rusu\etal\cite{PNN} introduce \emph{Progressive Neural Networks} (PNN), an effective approach for learning in a sequential multi-task setting. In PNN, a new network and lateral connections for each additional task are added in order to enable knowledge transfer, which speeds up the training of subsequent tasks. The additional network parts let the architecture grow super-linearly, while our network scales economically with an increasing number of tasks.
Another strong approach is introduced by Teh\etal\cite{Distral}. Their algorithm, \emph{Distral}, learns multiple tasks at once by sharing knowledge through a distillation process of an additional shared policy network. In contrast to our approach, this requires an aligned action space and a separate network for each task. We compare against Distral and PNN in our experiments.

\section{Background}

\begin{figure*}[t]
	\centering
    \includegraphics[width=\textwidth]{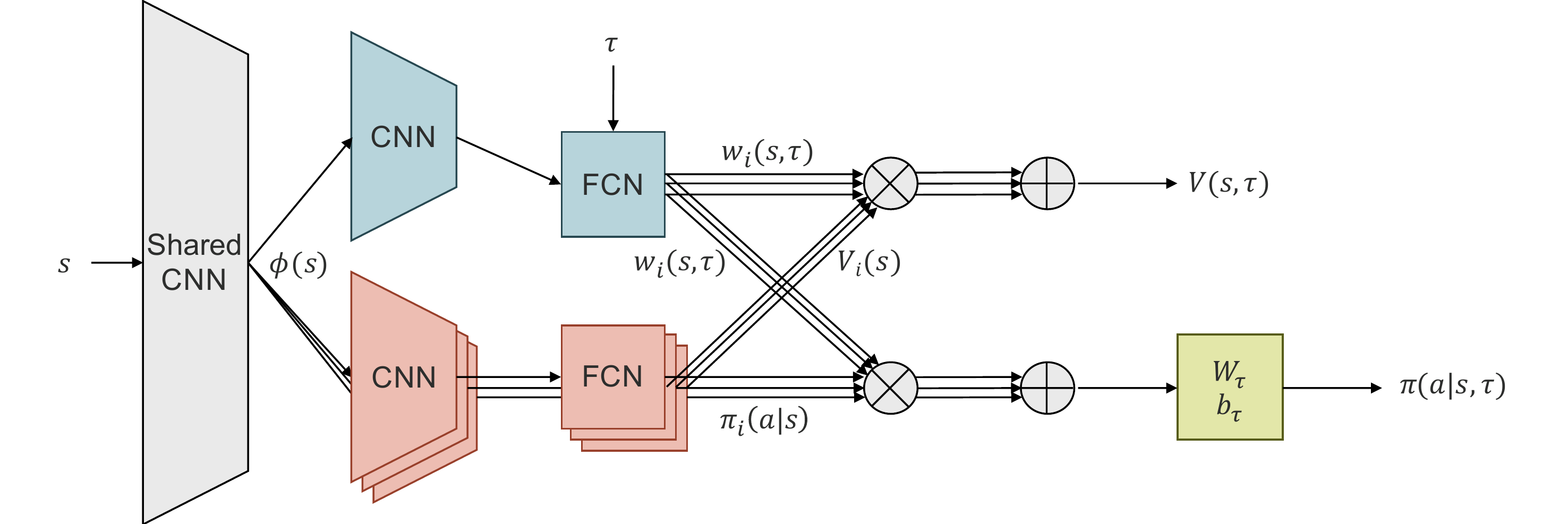}
	\caption{\label{fig:model-architecture} Our architecture consists of an attention network (blue), which decides the weighting of the sub-network outputs (red) to generate the policy and value function on a task and state basis. The first convolutional layers are shared between the attention network and the sub-networks. The weighted sub-network policies are transformed by a task-specific layer (in green) to account for different numbers of possible actions in different tasks.}
\end{figure*}

In reinforcement learning, an agent learns through interactions with an environment. The agent repeatedly chooses an action $a_t\in \mathcal{A}$
at step $t$ and observes a reward $r_t\in \mathbb{R}$ and the next state $s_{t+1}\in \mathcal{S}$, where $\mathcal{A}$ and $\mathcal{S}$ denote the sets of possible actions and states, respectively. The agent chooses the actions according to a policy $\pi(a_t|s_t): S\times A \rightarrow [0,1]$
which indicates the probability of choosing action $a_t$ in state $s_t$. The objective is to find a policy that maximizes the expected discounted return, i.e., to find
\[\pi^* = \max_\pi \left(\mathbb{E}_\pi\left[\sum_{t'=0}^\infty \gamma^{t'} r_{t'}\right]\right)\]
where $\gamma \in [0,1]$ is the discount factor for future rewards.

In this work, we train on this objective using asynchronous advantage actor-critic training~(A3C, \cite{A3C}), a well established policy gradient method that uses multiple asynchronous actors for experience collection. However, our approach is general and can be readily applied to most on- and off-policy deep reinforcement learning algorithms.

In multi-task reinforcement learning, the goal is to solve a set of tasks $\mathcal{T}$ simultaneously by training a policy $\pi(a_t|s_t,\tau)$ and value function $V(s_t,\tau)$, also referred to as critic, for each task $\tau\in\mathcal{T}$. While the objective to maximize the discounted rewards in each of the tasks remains unchanged, an additional goal is to share knowledge between tasks to accelerate training.

\section{Architecture} \label{sec:architecture}

Our network architecture, as shown in Figure~\ref{fig:model-architecture}, consists of a number of independent sub-networks and an attention module that weights the output of all sub-networks to generate a weighted policy and value function per task. The policies are then used to choose the next action in each of the environments. The attention and sub-networks all operate on top of a shared CNN that extracts high-level features of the environments. 
The attention network determines whether sub-networks become specialized on certain tasks, or whether they learn features that are shared across a group of tasks. However, we do not explicitly enforce this. Thus, we do not require any a-priori knowledge about the nature of the tasks or about their similarity. In other words, we do not make any assumptions about whether potential for positive or negative transfer exists.

\subsection{Shared Feature Extractor}

The first stage of our architecture consists of a CNN that outputs a state-embedding $\phi(s)$. The embedding $\phi(s)$ is shared among all following sub-networks as well as the attention network. Thus, $\phi(s)$ will learn general high-level features that are relevant for all subsequent parts of the architecture. Since we do not decrease the dimensionality of the input in these layers, the architecture can in the (worst) case, where no information can be shared, learn an approximate identity mapping from $s$ to $\phi(s)$ and leave the specialization to the sub-networks.

\subsection{Attention Network}

One could think of several ways how to combine the different sub-network outputs into a policy per task. One way would be to choose in each time step one of the sub-networks directly as policy. However, this sort of hard attention leads to noisy gradients (since a stochastic sampling operation would be added to the computation graph) and no complex interactions of several sub-networks could be learned. Therefore we employ a soft attention mechanism, where the final output is a linear combination of the sub-networks' outputs.
Intuitively, this allows all sub-networks that are helpful to contribute to the policy and value function. This can also be seen as an ensemble, where different sub-networks with possibly different specializations vote on the next action, but where the final decision is governed by an attention network. 

More concretely, the attention network consists of a CNN that operates on the shared embedding $\phi(s)$. The output of the CNN is fed into a fully connected network (FCN) that projects the output into a latent vector. This vector is then concatenated with a one-hot encoding of the task ID $\tau$ from which the input $s$ originates, and processed further in the fully connected network. Finally, a linear layer with softmax activation produces the attention weights $w_i(s, \tau)$, which decide the contribution of the policy and value functions of each sub-network $i$ in state $s$ of task $\tau$.

\subsection{Sub-Networks}

We use $N$ sub-networks that contribute to the final weighted policy and value function. The number of sub-networks can be chosen based on resource requirements and/or availability. In a practical application of our method, one would choose the maximum number of networks for which the entire model still fits into memory. Unused sub-networks can be automatically ignored by the attention network (see Section~\ref{subsec:analyzeattweights}), and could potentially be pruned to reduce the overall number of parameters. In our experiments we choose a small number of networks to show that we can achieve comparable or superior performance to state of the art methods while requiring substantially fewer parameters. Specifically, we chose the number of sub-networks $N$ depending on the number of tasks. That is, we roughly add one sub-network for four tasks. More precisely we let $N = \lfloor(|\mathcal{T}| + 2) / 4\rfloor + 1$, as we found this scaling to work well in our experiments. The sub-networks can act independently, as in an ensemble, or specialize on certain types of (sub-)tasks. The exact mode of operation depends on the nature of the tasks and is governed by the attention network. In other words, if the attention network decides that specialization is most beneficial, then the sub-networks will be encouraged to specialize, and vice versa. 

The sub-networks all have the same architecture and get the embedding $\phi(s)$ as input. First, a CNN learns to extract sub-network specific features from $\phi(s)$ that are then passed to a FCN. From the last hidden representation of the FCN, a linear layer directly outputs the value function estimate $V_i(s)$ for the $i$-th sub-network. A softmax layer maps the last hidden representation of the FCN to a $|\mathcal{A}_{max}|$-dimensional vector $pi_i(a|s)$, where $|\mathcal{A}_{max}|$ is the largest action space size across all tasks.

\subsection{Attentive Multi-Task Network}
The attention weighted $\pi_i(a|s)$ is in the end fed to a task-specific linear layer that maps it to the action dimension of each task, and a final softmax normalization is applied to generate a valid probability distribution over actions, i.e., a policy. More formally, the sub-network outputs $\pi_i(a|s)$ are combined into the final policy as 

\[\pi(a|s,\tau) = \text{softmax}\left(W_\tau\cdot\left( \sum_i^N{\pi_i(a|s) w_i(s,\tau)}\right) + b_\tau\right)\]

where $W_\tau\in\mathbb{R}^{|\mathcal{A}_\tau|\times |\mathcal{A}_{max}|}$ is a task-specific weight matrix and $b_\tau\in\mathbb{R}^{|\mathcal{A}_\tau|}$ is a task-specific bias. Note that $W_\tau$ and $b_\tau$ are shared across the sub-networks and only depend on the task.

Putting everything together, we use the attention weights $w_i(s,\tau)$ to also compute the final value function $V(s,\tau)$ from the outputs of the sub-networks as 

\[V(s,\tau)=\sum_{i=1}^N w_i(s,\tau) V_{i}(s)\]

\begin{figure}[t]
\centering
\begin{subfigure}[t]{.3\textwidth}
	\centering
    \includegraphics[width=1\textwidth]{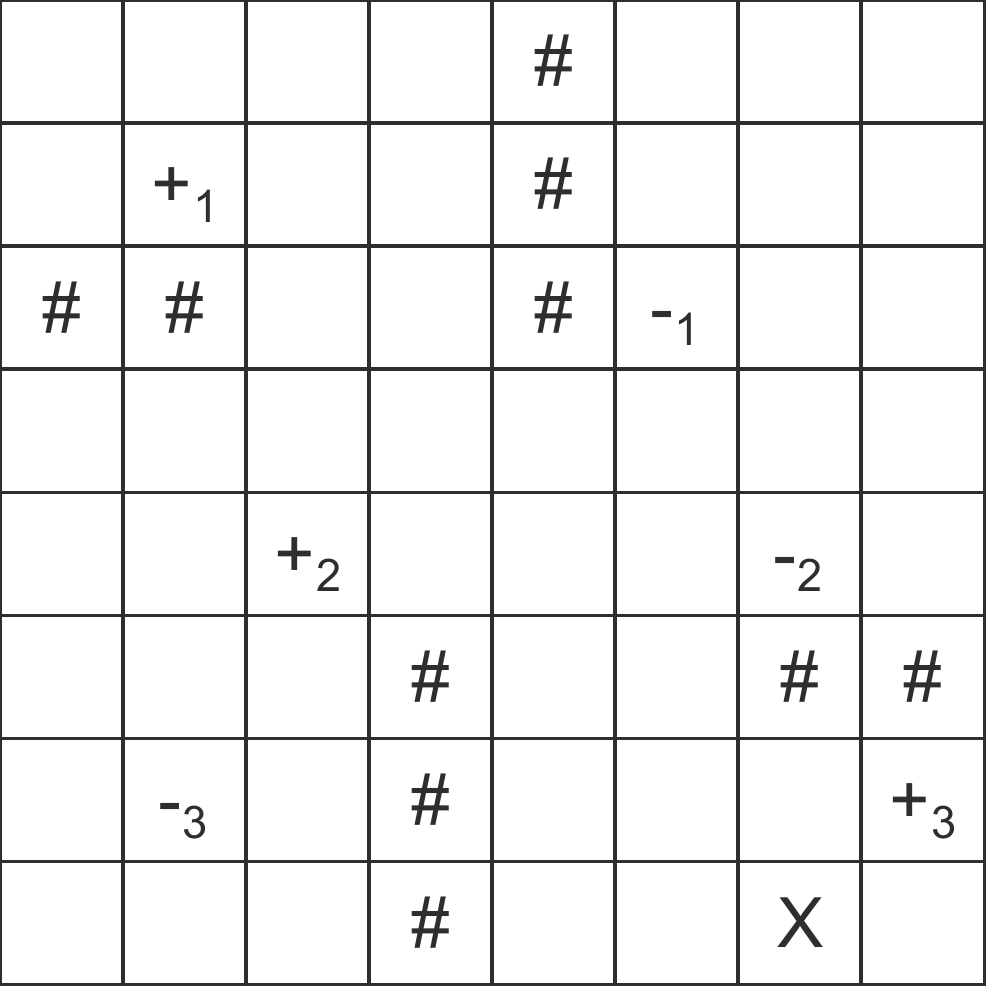}
	\caption{\label{fig:tasks} $\#$ are walls and $X$ is the target. $+$ ($-$) denote objects giving positive (negative) rewards. If only one bonus (penalty) object is present, it is located at $+_1$ ($-_1$).}
\end{subfigure}
\hfill
\begin{subfigure}[t]{.3\textwidth}
	\centering
    \includegraphics[width=1\textwidth]{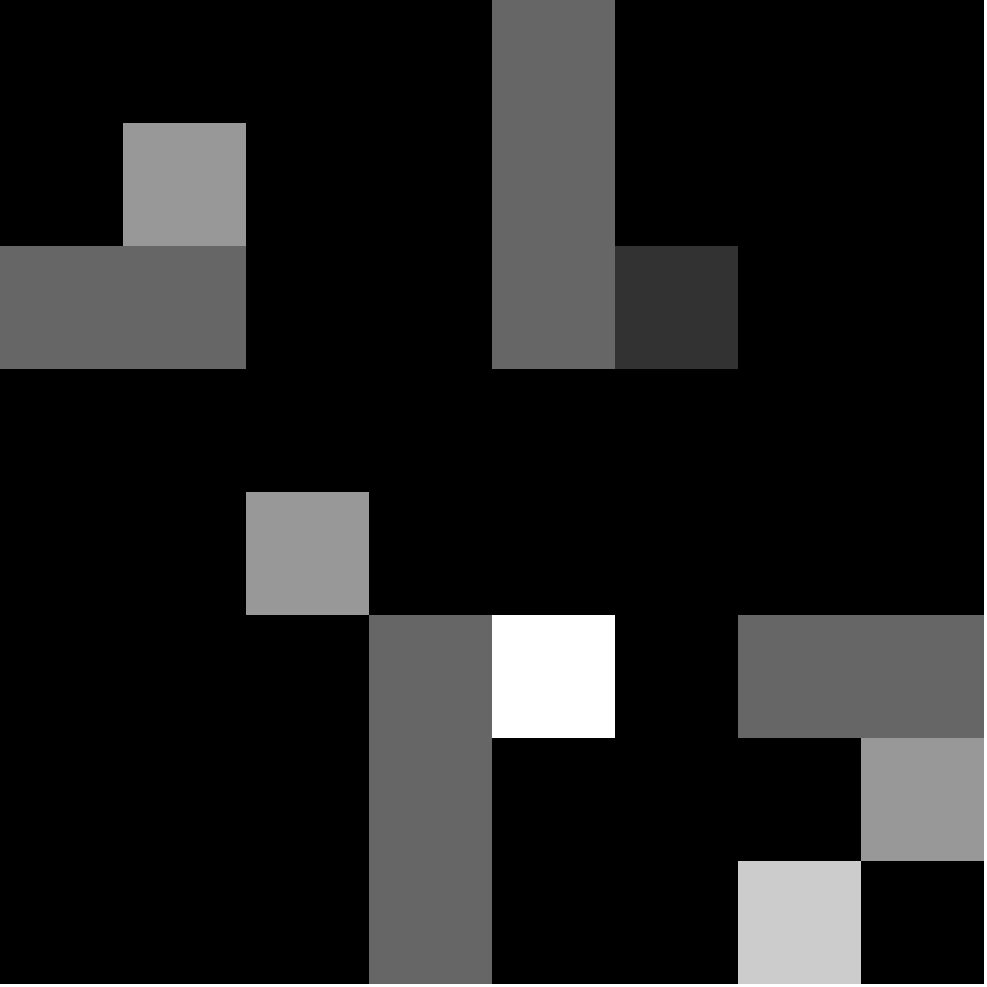}
	\caption{\label{fig:tasks-image} A state space of a grid world task. The player is the white square. Walls, the target, three bonus objects and one penalty object are present.}
\end{subfigure}
\hfill
\begin{subfigure}[t]{.3\textwidth}
	\centering
    \includegraphics[width=1\textwidth]{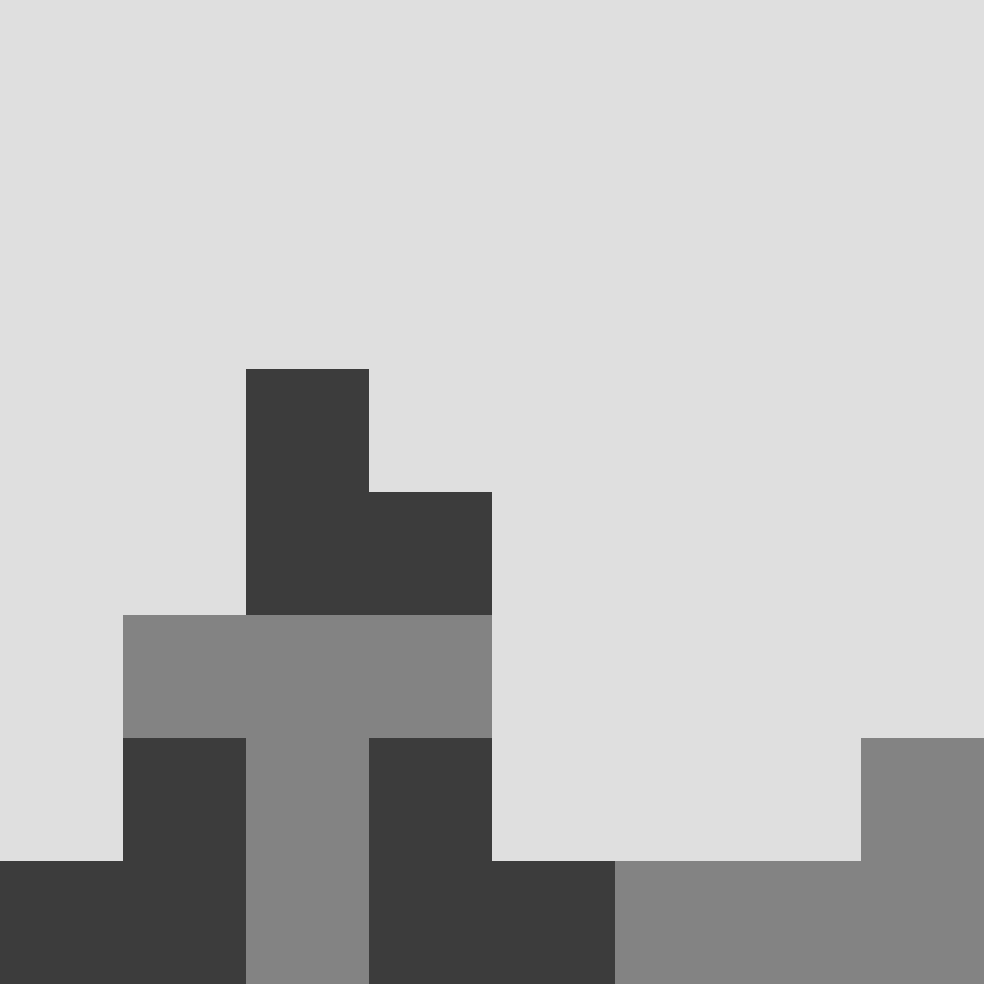}
	\caption{\label{fig:connect-four-image} Example state of the connect four task. The agent controls the dark tokens, the random opponent controls the gray tokens.}
\end{subfigure}
\caption{(a) Grid world environment template.  (b) Example of a concrete grid world instantiation. (c) Example of a concrete connect four state.
}
\end{figure}

\section{Task Environments}\label{section:environments}

To evaluate our approach we create a set of environments which are designed to have the potential for positive as well as negative knowledge transfer. Since we aim at evaluating our approach on a large set of tasks, we opt for simple, easy to generate environments, even though our initial results on the Arcade Learning Environment \cite{machado17arcade} (not reported here) were promising as well. We leave the adaption of our methodology to more complex environments to future work as we aim to show the evolution of transfer depending on the number of tasks in this report, which was not feasible on more complex tasks within our resource constraints due to the large amount of networks trained (600 for Figure~\ref{fig:steps} alone) and experiments conducted.

\subsection{Grid Worlds}\label{section:grid_worlds}

The first set of environments contains $20$ grid world tasks. The environments of this set consist of $8 \times 8$ gray-scale images representing the state of the environment. The agent is a single pixel in the grid and the possible actions are moving up, down, left or right. For all tasks, the goal is to reach a target pixel where a positive reward is received and the episode terminates. The environments can also contain additional objects that represent positive/negative rewards, as well as impassable walls.
All objects in the environments are at fixed locations, and only the starting location of the player is random. Figure~\ref{fig:tasks} shows the template for all tasks and Figure~\ref{fig:tasks-image} shows an example of such an environment as seen by the agent.

In the following we give a detailed description of every variation, each defining a task.
In the first task, the goal is to find the target as fast as possible. No walls or additional rewards are put into the environment, just the agent and the target. To encourage speed, the agent is penalized with a small negative reward at every step. In the other tasks there is no such penalty, but if the player leaves the board, a negative reward of $-0.5$ is observed and the episode is terminated. This is also the only difference between task one and two: The goal of the second task is to reach the target without leaving the board. 
In the third task, a bonus object is added at location marked $+_1$ in Figure~\ref{fig:tasks} that yields a positive reward when collected. 
In the fourth task, additionally to the bonus object of task $3$, another bonus object at location $+_2$ and a penalty object at the location marked with $-_1$ are added. The penalty object yields a negative reward when collected. 
The fifth and sixth task both contain three bonus objects at the locations marked with a $+$ in Figure~\ref{fig:tasks}, where the sixth task additionally contains another penalty object at location $-_2$. 
Tasks $7$ and $8$ are visually indistinguishable from tasks $4$ and $5$, but we invert the rewards of the bonus and penalty objects in order to test negative transfer. Similar to these two tasks, tasks $9$ and $10$ consist of three objects looking like penalty objects (at locations marked with $-$) but yielding positive reward. Task $10$ additionally contains an object that looks like a bonus object at location $+_1$ (see Figure~\ref{fig:tasks}) yielding a negative reward.
Tasks $11$ to $20$ are the same as tasks $1$ to $10$ but additionally contain impassable walls. 
The maximum achievable reward is set to $1.0$ for all tasks, distributed equally among bonus objects and target. For example, if there are three bonus objects, the target and bonus objects yield rewards of $0.25$ each. 
The penalty objects give a negative reward that is equal in magnitude to the bonus objects' positive reward.
In addition, walking into a wall yields a reward of $-0.5$.
Furthermore, if the agent does not reach the target after $200$ steps, the task terminates without any additional reward.

\subsection{Connect Four}\label{section:connect-four}

To test the behavior of our model on unrelated tasks with little to no potential for knowledge transfer, we generate environments from a completely different domain. We implement a two-player game based on connect four. Each location or token is represented by a single pixel. The agent drops in a token from the top, which then appears on top of the top most token in the column, or in the bottom row if the column was empty. The goal of this task is to have four tokens in a horizontal, vertical or diagonal line. Our connect four tasks consist of $8$ rows and $8$ columns, and thus looks visually similar to the grid world tasks, but has otherwise no relation to them. The agent has $8$ different actions to choose from, indicating in which column the token is to be dropped. An example of this is shown in Figure~\ref{fig:connect-four-image}.
If the agent plays an invalid action, i.e., if the chosen column is already full, the agent loses the game immediately. When the agent wins the game it receives a reward of $1$, and $-1$ if it loses. In case of a tie the reward is $0$. The opponent chooses a \emph{valid} action uniformly at random.
We additionally implement three variations of this basic connect four task. The goal of the first variation is to connect five tokens instead of four. The second and third variation rotate the state of the connect four and connect five tasks by $90$ degrees, such that the players now choose rows and not columns.

\section{Experiments and Results} \label{sec:expresults}

We evaluate the performance of our architecture on the set of grid worlds described before and compare the results to two state of the art architectures: Progressive Neural Networks (PNN) \cite{PNN} and Distral \cite{Distral}.
PNN learns tasks sequentially by freezing already trained networks and adding an additional network for each new task. The new networks are connected to previous ones to allow knowledge transfer. The order in which the tasks are trained with our PNN implementation is sampled randomly for all experiments. 
In contrast to PNN, but similar to our approach, Distral learns all tasks simultaneously. Here, a distilled policy $\hat\pi_0$ is used for sharing and transferring knowledge, while each task also has its own network to learn task-specific policies $\hat\pi_\tau$. We implement the \textit{KL+ent 2col}
approach (see \cite{Distral}). The distilled policy network and the task-specific networks have the same network architecture as the base PNN model which is listed as \emph{Base network} in Table~\ref{table:architectures}.
\begin{table}[t]
\centering
\caption{Architecture details for the policy networks (value output is omitted for readability). The base network is the basic network building block for Distral and PNN, each having one such base network per task. Additionally PNN has lateral connections and Distral has an additional base network for the shared policy. The columns \emph{Shared CNN}, \emph{Sub-networks} and \emph{Attention network} describe our architecture (see Section~\ref{sec:architecture} and Figure~\ref{fig:model-architecture}). The $+$ in the attention network Layer 4 indicates concatenation of the task embedding.}
\begin{tabular}{|l|l|l|l|l|}
\hline
& Base network & Shared CNN & Sub-networks & Attention network \\
\hline
Layer 1 & 3x3x16, stride 2 & 3x3x32, stride 2 & - & - \\
Layer 2 & 3x3x16, stride 1 & 3x3x32, stride 1 & - & - \\
Layer 3 & 3x3x16, stride 1 & - & 3x3x16, stride 1 & 3x3x16, stride 1 \\
Layer 4 & FC 256 & - & FC 256 & FC $N\cdot|\mathcal{T}| + |\mathcal{T}|$\\
Layer 5 & Softmax $|\mathcal{A}_\tau|$ & - & Softmax $|\mathcal{A}_{max} |$ & FC 256 \\
Layer 6 & - & - & Softmax $|\mathcal{A}_\tau|$ & Softmax $N$ \\
\hline
\end{tabular}
\label{table:architectures}
\end{table}
Note that even though our architecture starts with more filters in the shared CNN when compared to the base architecture, this does not give us a parameter advantage, since those filters are shared across all tasks while Distral and PNN get additional CNN parameters for each additional task.
For all approaches and all experiments, we use the same hyper parameters which are summarized in Table~\ref{table:hyperparameters}. We chose these hyper parameters based on the performance of all three approaches on multiple grid world tasks such that no approach has an unfair advantage.
We use the smallest multiple of $|\mathcal{T}|$ (the number of tasks) which is equal or larger than $24$ for the number of parallel workers in the A3C training and distribute tasks equally over the workers. The loss function is minimized with the Adam optimizer~\cite{ADAM_optimizer}.
For PNN we had to reduce the number of workers to 16, as the memory consumption for a large number of tasks was too high. 
For Distral, we set $\alpha=0.5$ and $\beta=10^4$ and compute the policy as
$$\hat\pi_i(a|s) = \text{softmax}(\alpha h(a|s) + f(a|s))$$
where $h$ is the output of the distilled network and $f$ is the $\beta$-scaled output of the task-specific network (see Appendix B.2 of \cite{Distral}).

\begin{table}[t]
\centering
\caption{Hyper parameters used for the experiments.}
\begin{tabular}{|ll|ll|}
\hline
Stacked input frames: & 1 & Discount factor $\gamma$: & 0.99 \\
Adam learning rate: & 1e-4 & Rollout length: & 5 \\
Adam $\beta_1$: & 0.9 & Entropy regularization: & 0.02 \\
Adam $\beta_2$: & 0.999 & Distral $\alpha$: & 0.5\\
Adam $\epsilon$: & 1e-08 & Distral $\beta$: & $10^4$\\
\hline
\end{tabular}
\label{table:hyperparameters}
\end{table}

\subsection{Model Size}

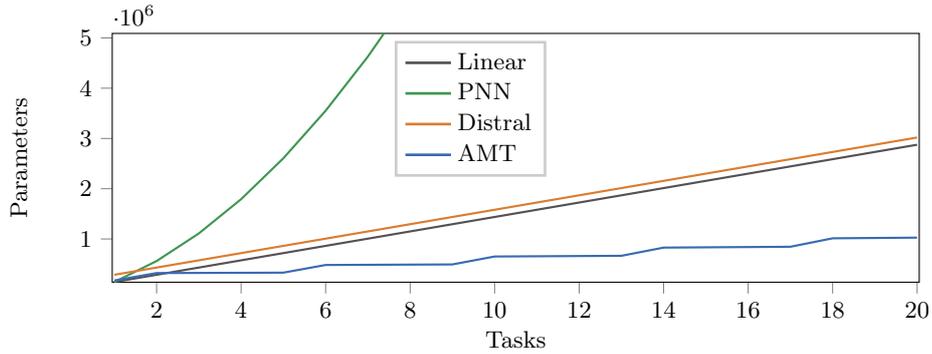
\begin{figure}[t]
	\centering
\begin{tikzpicture}

\definecolor{color0}{rgb}{0.325490196078431,0.317647058823529,0.329411764705882}
\definecolor{color1}{rgb}{0.243137254901961,0.588235294117647,0.317647058823529}
\definecolor{color2}{rgb}{0.854901960784314,0.486274509803922,0.188235294117647}
\definecolor{color3}{rgb}{0.223529411764706,0.415686274509804,0.694117647058824}

\begin{axis}[
xlabel={Tasks},
ylabel={Parameters},
xmin=0.95, xmax=20.05,
ymin=140000, ymax=5089585.8953066,
tick align=outside,
tick pos=left,
ytick={0,1000000,2000000,3000000,4000000,5000000},
scaled y ticks=base 10:-6,
height=\plotHeight,
width=\linewidth,
x grid style={white!69.01960784313725!black},
y grid style={white!69.01960784313725!black},
legend entries={{Linear},{PNN},{Distral},{AMT}},
legend style={at={(0.35,0.97)}, anchor=north west, draw=white!80.0!black, font=\small, line width=1pt},
legend cell align={left}
]
\addlegendimage{no markers, color0}
\addlegendimage{no markers, color1}
\addlegendimage{no markers, color2}
\addlegendimage{no markers, color3}
\addplot [thick, color0]
table {%
1 143754
2 287508
3 431262
4 575016
5 718770
6 862524
7 1006278
8 1150032
9 1293786
10 1437540
11 1581294
12 1725048
13 1868802
14 2012556
15 2156310
16 2300064
17 2443818
18 2587572
19 2731326
20 2875080
};
\addplot [thick, color1]
table {%
1 143754
2 561050
3 1110442
4 1791930
5 2605514
6 3551194
7 4628970
8 5838842
9 7180810
10 8654874
11 10261034
12 11999290
13 13869642
14 15872090
15 18006634
16 20273274
17 22672010
18 25202842
19 27865770
20 30660794
};
\addplot [thick, color2]
table {%
1 287508
2 431262
3 575016
4 718770
5 862524
6 1006278
7 1150032
8 1293786
9 1437540
10 1581294
11 1725048
12 1868802
13 2012556
14 2156310
15 2300064
16 2443818
17 2587572
18 2731326
19 2875080
20 3018834
};
\addplot [thick, color3]
table {%
1 174390
2 321936
3 324540
4 327144
5 329748
6 482424
7 486054
8 489684
9 493314
10 651120
11 655776
12 660432
13 665088
14 828024
15 833706
16 839388
17 845070
18 1013136
19 1019844
20 1026552
};
\end{axis}

\end{tikzpicture}
	\caption{\label{fig:parameters} Number of parameters of the different architecture choices with an increasing number of tasks. \emph{Linear} represents training each task in a separate network of the size of PNNs base model.}
\end{figure}

First, we compare the model sizes of our Attentive Multi-Task (AMT) architecture, PNN, Distral and \emph{Linear}. Linear simply represents training a separate network (same size as the base network) on each task which leads to a linear increase in parameters with each additional task. The results are shown in Figure~\ref{fig:parameters}. 
In our experiments we add a new sub-network to AMT for every fourth task, thus the number of network parameters grows more slowly with the number of tasks than in the other approaches. Depending on memory requirements we can easily increase or decrease the total number of parameters since we do not assign sub-networks to tasks a-priori; more difficult tasks can automatically be assigned more effective network capacity by the attention network. 

Distral uses slightly more parameters than having a separate network for each task due to the additional distilled policy network. The only way to reduce the number of total parameters would be to decrease the size of the task networks. However, unlike our approach, doing so could more strongly affect difficult tasks that require more network capacity to be solved, or tasks that cannot profit from the distilled policy due to a lack of transfer potential. One could tune each task network individually and, e.g., use larger networks for more difficult tasks, but this would require a substantial tuning effort. In contrast, our method assigns effective network capacity automatically, and can thus utilize the available network parameters more efficiently.

PNN also adds a new sub-network for each task and additionally connects all existing sub-networks to the newly added one. Thus, the number of total parameters grows super-linearly in the number of tasks. This parameter explosion causes high memory consumption and high computational costs, which can quickly become a problem when training on an increasing number of tasks with limited hardware.

\subsection{Sample Efficiency vs. Number of Tasks}

\begin{figure}[t]
	\centering
\begin{tikzpicture}

\definecolor{color0}{rgb}{0.243137254901961,0.588235294117647,0.317647058823529}
\definecolor{color1}{rgb}{0.854901960784314,0.486274509803922,0.188235294117647}
\definecolor{color2}{rgb}{0.223529411764706,0.415686274509804,0.694117647058824}
\definecolor{color3}{rgb}{0.325490196078431,0.317647058823529,0.329411764705882}

\begin{axis}[
xlabel={Tasks},
ylabel={Steps},
xmin=0.95, xmax=20.05,
ymin=700000, ymax=17025000,
tick align=outside,
tick pos=left,
scaled y ticks=base 10:-6,
height=0.95*\plotHeight,
width=\linewidth,
x grid style={white!69.01960784313725!black},
y grid style={white!69.01960784313725!black},
legend entries={{Linear},{PNN},{Distral},{AMT}},
legend style={at={(0.03,0.97)}, anchor=north west, draw=white!80.0!black,line width=1pt},
legend cell align={left}
]
\addlegendimage{no markers, color3}
\addlegendimage{no markers, color0}
\addlegendimage{no markers, color1}
\addlegendimage{no markers, color2}
\path [draw=color0, fill=color0, opacity=0.2] (axis cs:1,900000)
--(axis cs:1,500000)
--(axis cs:2,1400000)
--(axis cs:3,2000000)
--(axis cs:4,2500000)
--(axis cs:5,2800000)
--(axis cs:6,3400000)
--(axis cs:7,3700000)
--(axis cs:8,4400000)
--(axis cs:9,4800000)
--(axis cs:10,6100000)
--(axis cs:11,6300000)
--(axis cs:12,6500000)
--(axis cs:13,6800000)
--(axis cs:14,7100000)
--(axis cs:15,8100000)
--(axis cs:16,9000000)
--(axis cs:17,9200000)
--(axis cs:18,9900000)
--(axis cs:19,10500000)
--(axis cs:20,10700000)
--(axis cs:20,11600000)
--(axis cs:20,11600000)
--(axis cs:19,11300000)
--(axis cs:18,11000000)
--(axis cs:17,10800000)
--(axis cs:16,10500000)
--(axis cs:15,10000000)
--(axis cs:14,9400000)
--(axis cs:13,9100000)
--(axis cs:12,8400000)
--(axis cs:11,7800000)
--(axis cs:10,7400000)
--(axis cs:9,6500000)
--(axis cs:8,6300000)
--(axis cs:7,5700000)
--(axis cs:6,5400000)
--(axis cs:5,4500000)
--(axis cs:4,4000000)
--(axis cs:3,3800000)
--(axis cs:2,2700000)
--(axis cs:1,900000)
--cycle;

\path [draw=color1, fill=color1, opacity=0.2] (axis cs:1,900000)
--(axis cs:1,500000)
--(axis cs:2,1000000)
--(axis cs:3,1600000)
--(axis cs:4,1000000)
--(axis cs:5,2000000)
--(axis cs:6,2600000)
--(axis cs:7,3600000)
--(axis cs:8,2400000)
--(axis cs:9,3100000)
--(axis cs:10,4300000)
--(axis cs:11,3900000)
--(axis cs:12,3700000)
--(axis cs:13,4800000)
--(axis cs:14,4900000)
--(axis cs:15,4700000)
--(axis cs:16,5100000)
--(axis cs:17,6100000)
--(axis cs:18,6500000)
--(axis cs:19,6500000)
--(axis cs:20,6900000)
--(axis cs:20,7800000)
--(axis cs:20,7800000)
--(axis cs:19,7200000)
--(axis cs:18,6900000)
--(axis cs:17,7000000)
--(axis cs:16,5900000)
--(axis cs:15,5800000)
--(axis cs:14,5500000)
--(axis cs:13,5700000)
--(axis cs:12,5600000)
--(axis cs:11,4800000)
--(axis cs:10,4700000)
--(axis cs:9,3700000)
--(axis cs:8,3400000)
--(axis cs:7,3800000)
--(axis cs:6,3300000)
--(axis cs:5,2600000)
--(axis cs:4,1600000)
--(axis cs:3,1600000)
--(axis cs:2,1300000)
--(axis cs:1,900000)
--cycle;

\path [draw=color2, fill=color2, opacity=0.2] (axis cs:1,1300000)
--(axis cs:1,900000)
--(axis cs:2,2300000)
--(axis cs:3,2800000)
--(axis cs:4,3500000)
--(axis cs:5,3800000)
--(axis cs:6,5500000)
--(axis cs:7,4000000)
--(axis cs:8,6100000)
--(axis cs:9,6900000)
--(axis cs:10,7400000)
--(axis cs:11,6700000)
--(axis cs:12,7300000)
--(axis cs:13,7400000)
--(axis cs:14,7500000)
--(axis cs:15,7400000)
--(axis cs:16,8300000)
--(axis cs:17,8000000)
--(axis cs:18,7900000)
--(axis cs:19,8500000)
--(axis cs:20,8000000)
--(axis cs:20,9200000)
--(axis cs:20,9200000)
--(axis cs:19,9500000)
--(axis cs:18,9500000)
--(axis cs:17,10300000)
--(axis cs:16,8500000)
--(axis cs:15,8900000)
--(axis cs:14,8900000)
--(axis cs:13,10100000)
--(axis cs:12,9300000)
--(axis cs:11,8100000)
--(axis cs:10,7900000)
--(axis cs:9,7200000)
--(axis cs:8,7300000)
--(axis cs:7,6400000)
--(axis cs:6,6700000)
--(axis cs:5,5200000)
--(axis cs:4,4600000)
--(axis cs:3,3400000)
--(axis cs:2,3300000)
--(axis cs:1,1300000)
--cycle;

\addplot [thick, color3]
table {%
1 850000
2 1700000
3 2550000
4 3400000
5 4250000
6 5100000
7 5950000
8 6800000
9 7650000
10 8500000
11 9350000
12 10200000
13 11050000
14 11900000
15 12750000
16 13600000
17 14450000
18 15300000
19 16150000
20 17000000
};
\addplot [thick, color0]
table {%
1 700000
2 1750000
3 2250000
4 2950000
5 3350000
6 3850000
7 4400000
8 5600000
9 6100000
10 6950000
11 7450000
12 7850000
13 8300000
14 8750000
15 9050000
16 9300000
17 9650000
18 10150000
19 10750000
20 11250000
};
\addplot [thick, color1]
table {%
1 700000
2 1150000
3 1600000
4 1350000
5 2250000
6 3200000
7 3700000
8 3300000
9 3500000
10 4550000
11 4500000
12 5500000
13 5250000
14 5200000
15 5300000
16 5800000
17 6750000
18 6700000
19 6950000
20 7050000
};
\addplot [thick, color2]
table {%
1 950000
2 2750000
3 3100000
4 3850000
5 4250000
6 6250000
7 5000000
8 6700000
9 7050000
10 7700000
11 7750000
12 8050000
13 9000000
14 7900000
15 8000000
16 8400000
17 8600000
18 8400000
19 8650000
20 8200000
};
\end{axis}

\end{tikzpicture}
	\caption{\label{fig:steps} The number of steps required when trained on a set of tasks to reach an average score of at least $0.9$ and a score of at least $0.8$ on each task separately over $10^5$ steps. The median over $10$ runs and for task set sizes between $1$ and $20$ is shown, where the tasks are sampled randomly from all grid worlds tasks. The shaded area represents $30\%$ to $70\%$ performance of the runs. The number of steps for \emph{Linear} is calculated by extrapolating the median of all runs from the other three approaches that are trained on one task only.}
\end{figure}
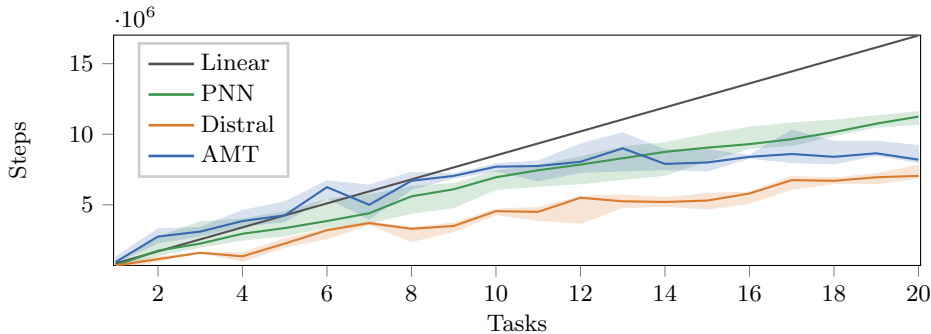

\begin{figure}[t]
	\centering
    \input{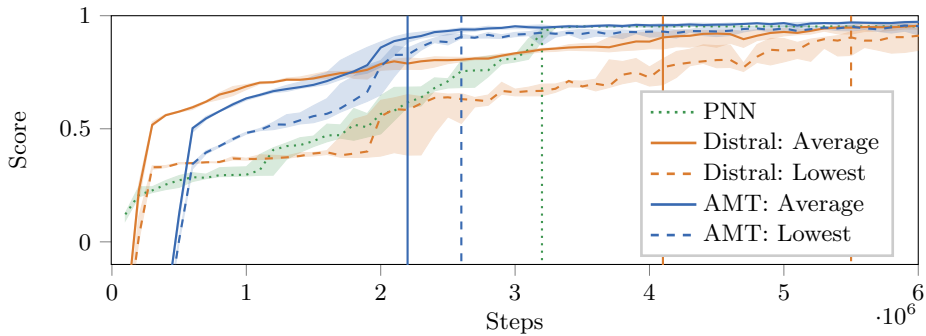}
	\caption{\label{fig:swap-actions} Median scores over $10$ runs. The shaded area represents $30\%$ to $70\%$ performance of the runs. Training is performed on $6$ tasks, where we switch the action dimensions for $3$ of them. This means that the agent goes to the left instead of the right, to the top instead of the bottom, and vice versa. For AMT and Distral, the solid line represents the average score of all tasks and the dashed line represents the task where the lowest score is observed. For PNN (dotted), a threshold score of $0.95$ has to be reached over $10^5$ steps before switching to the next task. For each approach, the vertical line is drawn at the first point where the score exceeds $0.9$ over $10^5$ steps.}
\end{figure}

In this section, we compare the performance of AMT to PNN and Distral when trained on an increasing number of tasks. We perform 10 runs for each approach and every number of tasks (from 1 to 20). For each of the 10 runs, the tasks are chosen uniformly at random without replacement from all $20$ grid world tasks. The tasks are considered solved if the average score over $10^5$ steps is at least $0.9$ and each individual task has a score of at least $0.8$. The results are shown in Figure~\ref{fig:steps}. 
The number of steps required to solve a given number of tasks scales sub-linearly for all three approaches, i.e., training on multiple tasks requires fewer interactions with the environment than training every task separately. This means that knowledge is shared between different tasks in all approaches as expected. 
For a larger amount of tasks, our approach is faster than PNN and only slightly worse than Distral in terms of steps required to reach the given performance threshold. Note however that our approach has substantially fewer parameters than the other approaches in this large number of tasks setup.
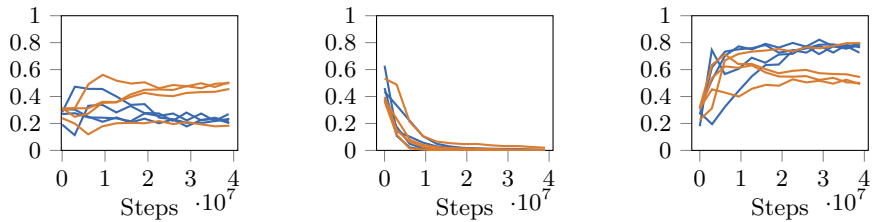
\begin{figure*}[t]
\centering
\begin{subfigure}[t]{.32\textwidth}
\begin{tikzpicture}

\definecolor{color0}{rgb}{0.223529411764706,0.415686274509804,0.694117647058824}
\definecolor{color1}{rgb}{0.854901960784314,0.486274509803922,0.188235294117647}

\begin{axis}[
xlabel={Steps},
xmin=-194866-1, xmax=40959325,
ymin=0, ymax=1,
width=\triplePlotWidth,
tick align=outside,
tick pos=left,
x grid style={white!69.01960784313725!black},
y grid style={white!69.01960784313725!black},
label style={font=\triplePlotAxesFontSize},
tick label style={font=\triplePlotTicksFontSize},
]
\addlegendimage{no markers, color0}
\addlegendimage{no markers, color1}
\addplot [thick, color0]
table {%
1702 0.282941908246338
2983180 0.473022206263109
6115946 0.456847919057115
9363599 0.45603464468561
12698465 0.39976227644718
15966951 0.337795313861635
19206229 0.343632612494996
22615823 0.243170157775772
25803233 0.231180254513436
29030756 0.204694579841217
32236688 0.202005677326959
35591251 0.21819828013031
38817288 0.269813817843756
};
\addplot [thick, color0, forget plot]
table {%
10313 0.191765252519587
2920676 0.113523402749889
6081389 0.331054040767026
9405779 0.337723779369785
12630932 0.281886181761181
15942493 0.329934138207309
19145587 0.277669160999594
22553928 0.274513321424656
25794366 0.219640699422975
28995913 0.280878834600819
32402869 0.223341768245138
35591664 0.242549384609445
38923705 0.214157667939693
};
\addplot [thick, color0, forget plot]
table {%
8926 0.306116046929601
2953511 0.302809919070715
6114656 0.249534456341555
9446258 0.212697085661042
12677514 0.244324254782193
15947913 0.206513735579769
19195350 0.235537608982594
22577087 0.201485113436779
25837979 0.229571040316498
29041114 0.177929091235857
32412354 0.226249635607983
35660211 0.234232528274222
38971924 0.202693618843468
};
\addplot [thick, color0, forget plot]
table {%
10024 0.269837431007563
2964246 0.275891417054215
6150820 0.244540577721452
9454366 0.242916222549287
12703319 0.237573946389842
15976434 0.216490639198156
19233149 0.274092244407485
22615472 0.256026852177345
25876721 0.273768220196174
29073028 0.222899799316881
32431099 0.273740340099668
35682699 0.217215083436269
38983840 0.234591102569834
};
\addplot [thick, color1]
table {%
6913 0.311580161554644
2964495 0.313555361345561
6174934 0.493147708550849
9475741 0.560772898823324
12709638 0.517398543731131
16049426 0.49655541115337
19247734 0.505407606411462
22597632 0.461382064253393
25857876 0.48544618487358
29089177 0.476122925076821
32429911 0.498448405903999
35594353 0.470798155575088
38953351 0.505591149583007
};
\addplot [thick, color1, forget plot]
table {%
8899 0.23939803400726
2935395 0.199382328776398
6078834 0.119058756402346
9371556 0.178462600316664
12633884 0.201204567282188
15930993 0.204926480360404
19093623 0.201380236394177
22625000 0.21753104309542
25835882 0.195834972604056
29087900 0.209992377595468
32510712 0.192580941349569
35669273 0.178797175261107
39008962 0.183422042339137
};
\addplot [thick, color1, forget plot]
table {%
8276 0.299900319239106
2961686 0.312657725329351
6107465 0.312749271892537
9367602 0.360410298437182
12652886 0.358353186586892
15922081 0.414721417833458
19180660 0.449658491545253
22572637 0.452600437809121
25819187 0.448278658317798
29032246 0.476912155785043
32418462 0.461333153562412
35676438 0.494727166839923
38995615 0.499306080543031
};
\addplot [thick, color1, forget plot]
table {%
6794 0.315121747026539
2949036 0.249349759835186
6098982 0.272965946432316
9370570 0.352762806445662
12638830 0.356619929725474
15919262 0.396097385522092
19177798 0.427330323692524
22593432 0.409297493521614
25824221 0.402466659443546
29011897 0.426575930582152
32414905 0.432756383461181
35648063 0.435218241780695
38978009 0.455922341858498
};
\end{axis}

\end{tikzpicture}
	\caption{\label{fig:attention-weights-1} First sub-network.}
\end{subfigure}
\hfill
\begin{subfigure}[t]{.32\textwidth}
\begin{tikzpicture}

\definecolor{color0}{rgb}{0.223529411764706,0.415686274509804,0.694117647058824}
\definecolor{color1}{rgb}{0.854901960784314,0.486274509803922,0.188235294117647}

\begin{axis}[
xlabel={Steps},
xmin=-1948661, xmax=40959325,
ymin=0, ymax=1,
tick align=outside,
tick pos=left,
width=\triplePlotWidth,
x grid style={white!69.01960784313725!black},
y grid style={white!69.01960784313725!black},
label style={font=\triplePlotAxesFontSize},
tick label style={font=\triplePlotTicksFontSize},
]
\addlegendimage{no markers, color0}
\addlegendimage{no markers, color1}
\addplot [thick, color0]
table {%
1702 0.436000970276919
2983180 0.332506001597704
6115946 0.219835211231251
9363599 0.104960158166259
12698465 0.0489131446288091
15966951 0.0288570626366018
19206229 0.0186631481323127
22615823 0.0156773059638313
25803233 0.0124453260713831
29030756 0.0109844454564154
32236688 0.0106183797487934
35591251 0.00680279382739708
38817288 0.0046820609893231
};
\addplot [thick, color0, forget plot]
table {%
10313 0.628659831755089
2920676 0.140857174014202
6081389 0.102450666253013
9405779 0.0569528563305588
12630932 0.0297444555155862
15942493 0.0197165708970088
19145587 0.0112018832240746
22553928 0.00837225159583159
25794366 0.00667379562968789
28995913 0.00441938929901833
32402869 0.00390073701485305
35591664 0.00241102423634129
38923705 0.0015337322168335
};
\addplot [thick, color0, forget plot]
table {%
8926 0.378562037089858
2953511 0.174331253902479
6114656 0.0484090616286857
9446258 0.0147605906464066
12677514 0.00590829575039512
15947913 0.00216474686641802
19195350 0.00146544183990615
22577087 0.000603727662315831
25837979 0.000359435008389482
29041114 8.41937613033694e-05
32412354 5.49843786367048e-05
35660211 3.90211971791397e-05
38971924 1.90444840368829e-05
};
\addplot [thick, color0, forget plot]
table {%
10024 0.463229833227216
2964246 0.107363506888199
6150820 0.0218297584803135
9454366 0.00652290503844275
12703319 0.00313661589119169
15976434 0.00127795704619197
19233149 0.000707179417240415
22615472 0.000364745917674335
25876721 0.00015040917026995
29073028 9.79780389106932e-05
32431099 7.31943972804492e-05
35682699 6.67693373609194e-05
38983840 1.32102628688281e-05
};
\addplot [thick, color1]
table {%
6913 0.374773768764554
2964495 0.233176359474056
6174934 0.078208114688445
9475741 0.039760897726272
12709638 0.0243970288923294
16049426 0.0168887245365315
19247734 0.0143728729553822
22597632 0.0109830900810359
25857876 0.0095886555265147
29089177 0.0070818349239275
32429911 0.00611847641200504
35594353 0.00433328747043781
38953351 0.00263589578610144
};
\addplot [thick, color1, forget plot]
table {%
8899 0.533179659133005
2935395 0.488684312863783
6078834 0.213086387876308
9371556 0.10570270869166
12633884 0.0655438144023367
15930993 0.0535933470184151
19093623 0.0467367112787083
22625000 0.0473150556737725
25835882 0.0378820646290828
29087900 0.0320277485544936
32510712 0.0319369100410531
35669273 0.0251683478779865
39008962 0.0199391350639288
};
\addplot [thick, color1, forget plot]
table {%
8276 0.397540789661986
2961686 0.149485133889348
6107465 0.0638423010405866
9367602 0.0261499490965196
12652886 0.0103848640703495
15922081 0.00543918249825227
19180660 0.00127490616322382
22572637 0.000382118352864784
25819187 0.000106072047856137
29032246 0.000104506357612266
32418462 4.98864107456334e-05
35676438 9.45043791388322e-06
38995615 1.00134815664582e-06
};
\addplot [thick, color1, forget plot]
table {%
6794 0.364455176423294
2949036 0.116160877651037
6098982 0.0145944899813547
9370570 0.00635667339865731
12638830 0.00123460042188094
15919262 4.5679439728092e-05
19177798 1.91645192260732e-05
22593432 4.11737657067194e-06
25824221 1.64659934344098e-06
29011897 6.29888664167418e-07
32414905 3.16235077294046e-07
35648063 1.27058673915217e-07
38978009 9.66775257838717e-08
};
\end{axis}

\end{tikzpicture}
	\caption{\label{fig:attention-weights-2} Second sub-network.}
\end{subfigure}
\hfill
\begin{subfigure}[t]{.32\textwidth}
\begin{tikzpicture}

\definecolor{color0}{rgb}{0.223529411764706,0.415686274509804,0.694117647058824}
\definecolor{color1}{rgb}{0.854901960784314,0.486274509803922,0.188235294117647}

\begin{axis}[
xlabel={Steps},
xmin=-1948661, xmax=40959325,
ymin=0, ymax=1,
tick align=outside,
tick pos=left,
width=\triplePlotWidth,
x grid style={white!69.01960784313725!black},
y grid style={white!69.01960784313725!black},
label style={font=\triplePlotAxesFontSize},
tick label style={font=\triplePlotTicksFontSize},
]
\addlegendimage{no markers, color0}
\addlegendimage{no markers, color1}
\addplot [thick, color0]
table {%
1702 0.281057112144701
2983180 0.194471792289705
6115946 0.32331686999017
9363599 0.439005195906202
12698465 0.551324582220319
15966951 0.633347630271123
19206229 0.637704249649705
22615823 0.741152535863159
25803233 0.756374423327002
29030756 0.784320973850529
32236688 0.787375949064182
35591251 0.774998926509389
38817288 0.725504124365133
};
\addplot [thick, color0, forget plot]
table {%
10313 0.179574913693348
2920676 0.745619416838944
6081389 0.566495290973117
9405779 0.605323364537434
12630932 0.688369364741979
15942493 0.650349291481282
19145587 0.711128957387335
22553928 0.717114427216746
25794366 0.773685506690821
28995913 0.714701779064431
32402869 0.77275749988327
35591664 0.755039591812574
38923705 0.784308599101173
};
\addplot [thick, color0, forget plot]
table {%
8926 0.315321915528992
2953511 0.522858823790694
6114656 0.702056486648742
9446258 0.772542327944007
12677514 0.749767449738535
15947913 0.791321518171268
19195350 0.762996946565017
22577087 0.797911148631211
25837979 0.770069518775652
29041114 0.821986715858472
32412354 0.773695381753373
35660211 0.765728450676694
38971924 0.797287338159302
};
\addplot [thick, color0, forget plot]
table {%
10024 0.26693273441057
2964246 0.616745082717954
6150820 0.733629659155583
9454366 0.750560872254404
12703319 0.75928943927842
15976434 0.782231394058645
19233149 0.725200571345558
22615472 0.743608392549282
25876721 0.726081369037238
29073028 0.777002224689257
32431099 0.726186472401929
35682699 0.782718156392962
38983840 0.76539569918179
};
\addplot [thick, color1]
table {%
6913 0.313646077808707
2964495 0.453268283846403
6174934 0.428644175319039
9475741 0.399466212317794
12709638 0.458204432558345
16049426 0.486555864187805
19247734 0.48021951285122
22597632 0.52763485232381
25857876 0.504965158912398
29089177 0.516795237354445
32429911 0.495433111706435
35594353 0.524868553485534
38953351 0.491772954291749
};
\addplot [thick, color1, forget plot]
table {%
8899 0.227422308891711
2935395 0.311933356929909
6078834 0.667854855608457
9371556 0.715834689862799
12633884 0.733251615004105
15930993 0.741480176948537
19093623 0.751883048150276
22625000 0.735153901155549
25835882 0.766282962578716
29087900 0.757979876935633
32510712 0.775482144620685
35669273 0.796034473060357
39008962 0.796638827369577
};
\addplot [thick, color1, forget plot]
table {%
8276 0.30255889200201
2961686 0.537857132126586
6107465 0.623408419782832
9367602 0.61343974537824
12652886 0.631261943436154
15922081 0.579839399229594
19180660 0.549066596653017
22572637 0.547017447479852
25819187 0.551615273637811
29032246 0.522983337084732
32418462 0.538616956636453
35676438 0.505263383361732
38995615 0.500692924674888
};
\addplot [thick, color1, forget plot]
table {%
6794 0.320423074141897
2949036 0.634489368910743
6098982 0.712439566700147
9370570 0.640880518818935
12638830 0.642145461565988
15919262 0.603856933198314
19177798 0.572650509018451
22593432 0.590698383036919
25824221 0.597531700717786
29011897 0.573423434191199
32414905 0.567243299673805
35648063 0.564781631069651
38978009 0.544077565664794
};
\end{axis}

\end{tikzpicture}
	\caption{\label{fig:attention-weights-3} Third sub-network.}
\end{subfigure}
\caption{\label{fig:attention-weights-grid-world} A smoothed average of the attention weights for $8$ different grid world tasks. Set 1 (blue) consists of tasks $3$ to $6$ and set 2 (orange) consists of tasks $7$ to $10$, for which the reward associated with the collectable objects are inverted. 
}
\end{figure*}

\subsection{Unaligned Action Spaces}


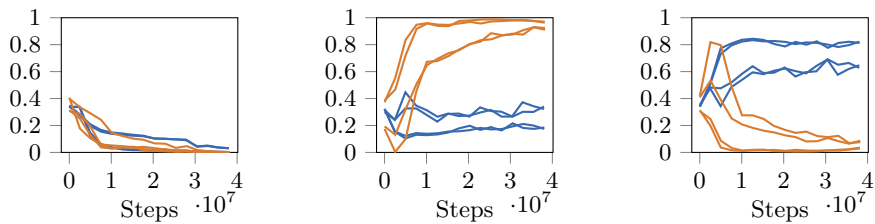
\begin{figure*}[t]
\centering
\begin{subfigure}{.32\textwidth}
\begin{tikzpicture}

\definecolor{color0}{rgb}{0.223529411764706,0.415686274509804,0.694117647058824}
\definecolor{color1}{rgb}{0.854901960784314,0.486274509803922,0.188235294117647}

\begin{axis}[
xlabel={Steps},
xmin=-1911728.9, xmax=40168526.9,
ymin=0, ymax=1,
width=\triplePlotWidth,
tick align=outside,
tick pos=left,
x grid style={white!69.01960784313725!black},
y grid style={white!69.01960784313725!black},
label style={font=\triplePlotAxesFontSize},
tick label style={font=\triplePlotTicksFontSize},
]
\addplot [thick, color0]
table {%
7481 0.341194130254514
2436497 0.276114580155623
4978934 0.197351738133214
7571552 0.153738809203861
10034867 0.138710696099684
12582158 0.131324263985711
15091545 0.12423410774632
17659633 0.120093042990475
20217858 0.103658601491138
22781692 0.102243514223532
25243546 0.0961306029272199
27785792 0.0954806284621508
30405588 0.0437126932947925
32887349 0.0509866771448142
35498132 0.0376602734128634
38028752 0.0305465283154538
};
\addplot [thick, color0, forget plot]
table {%
2008 0.348641701719977
2458066 0.281521260437339
4966947 0.209765655223769
7532012 0.166932152452493
10009771 0.148444376810633
12602705 0.140333841996964
15179732 0.130762631092408
17689136 0.123771021951629
20348512 0.102730816512397
22798967 0.100131973737117
25349559 0.0972504695557586
27865724 0.0903778982282887
30524536 0.0433621443431788
32975315 0.047673400585549
35487287 0.0353792622553731
38010775 0.0312566266329301
};
\addplot [thick, color0, forget plot]
table {%
12961 0.338711421899122
2463187 0.336594124484544
4971988 0.122885103685537
7555101 0.0597216949958088
10038890 0.0355987057296766
12643731 0.0226407684174112
15186414 0.016847073264547
17699103 0.0163739379360153
20312329 0.0153415760269678
22817317 0.0121098717197193
25378192 0.01232851570254
27897083 0.0124237619263544
30544357 0.00714521269614452
33024976 0.00546055500161294
35551853 0.00374026429022754
38087164 0.00260600413937268
};
\addplot [thick, color0, forget plot]
table {%
1010 0.336526460117764
2448678 0.339179408670676
4961599 0.146996748379685
7547284 0.0619659969733669
10033840 0.0388627631702656
12610777 0.0244842163766875
15181858 0.0205834940772014
17705900 0.0158929111520677
20357027 0.0138054082829816
22803337 0.0138712820404142
25365016 0.0122210069940511
27897270 0.0132033073456812
30542242 0.00612179736616776
33007214 0.00628389664846263
35528487 0.00414645479281492
38056723 0.00275556544390088
};
\addplot [thick, color1]
table {%
10918 0.307745707155478
2558265 0.28597013637273
5034129 0.191038637751281
7644862 0.0505540900760227
10190417 0.0287854184368341
12733012 0.0416107769453463
15351765 0.0417611347605483
17772025 0.0142130873758686
20447690 0.0119478878602765
22901585 0.00763516954494394
25457652 0.00540197349501523
28070789 0.00559239024813805
30517791 0.00342682814060933
33111900 0.00191311268700355
35623142 0.00118293357434486
38157358 0.000688443522761638
};
\addplot [thick, color1, forget plot]
table {%
2054 0.313777559935444
2533607 0.250889813809684
5001742 0.129217106435034
7603958 0.0369848309442249
10145082 0.0289364926742784
12692689 0.0357285947397801
15318813 0.0365119710037803
17743020 0.0244420367772832
20435472 0.0190767503832702
22896132 0.0238202590015577
25456273 0.0141150099013678
28068218 0.0118895064935916
30513616 0.0055462342011977
33134525 0.00253963293484178
35646702 0.00247983884059289
38179028 0.000716641998022018
};
\addplot [thick, color1, forget plot]
table {%
2251 0.40628180178729
2539428 0.176915198865563
4999591 0.104695748105043
7555108 0.058914011203204
10075345 0.0518674310484921
12614069 0.046063927688984
15257305 0.0364840963121617
17630105 0.041234949321458
20288897 0.0316023030553503
22851988 0.0222588869733643
25383598 0.00640986245737003
27980667 0.0103350633924658
30488995 0.00509851601567451
33046733 0.00213491427594545
35576767 0.000742800107797261
38250580 0.000204465535073624
};
\addplot [thick, color1, forget plot]
table {%
13754 0.398384387746002
2561434 0.33495711828723
5037573 0.290315366754628
7600637 0.240998508352222
10121106 0.145039714421287
12654046 0.122410384543014
15294726 0.101888879757337
17653529 0.094069886613976
20300292 0.0672174170236999
22859032 0.0645640021698042
25388128 0.0334024764785561
27981143 0.0463590760423679
30489372 0.0175937015958356
33037373 0.00910464201752814
35585195 0.00573941866744931
38255788 0.00183221797701074
};
\end{axis}

\end{tikzpicture}
	\caption{\label{fig:attention-weights-2-1} First sub-network.}
\end{subfigure}
\hfill
\begin{subfigure}{.32\textwidth}
\begin{tikzpicture}

\definecolor{color0}{rgb}{0.223529411764706,0.415686274509804,0.694117647058824}
\definecolor{color1}{rgb}{0.854901960784314,0.486274509803922,0.188235294117647}

\begin{axis}[
xlabel={Steps},
xmin=-1911728.9, xmax=40168526.9,
ymin=0, ymax=1,
width=\triplePlotWidth,
tick align=outside,
tick pos=left,
x grid style={white!69.01960784313725!black},
y grid style={white!69.01960784313725!black},
label style={font=\triplePlotAxesFontSize},
tick label style={font=\triplePlotTicksFontSize},
]
\addplot [thick, color0]
table {%
7481 0.315698844615859
2436497 0.241802524681194
4978934 0.326714622091992
7571552 0.326193411448839
10034867 0.278438852802436
12582158 0.228678198118524
15091545 0.285500175275222
17659633 0.28850323320784
20217858 0.266014950326409
22781692 0.334298695963262
25243546 0.307995864887477
27785792 0.302495186674559
30405588 0.267490552250305
32887349 0.372513374690005
35498132 0.342959104317495
38028752 0.322656942077564
};
\addplot [thick, color0, forget plot]
table {%
2008 0.299079126480853
2458066 0.242481355037954
4966947 0.446242341244913
7532012 0.347572800284722
10009771 0.313883032339314
12602705 0.265020391143002
15179732 0.288344676091894
17689136 0.269935163902118
20348512 0.27370474679158
22798967 0.296097375996967
25349559 0.31541618333678
27865724 0.266083983506306
30524536 0.265822392519131
32975315 0.301529162927917
35487287 0.300473367129782
38010775 0.3403865877125
};
\addplot [thick, color0, forget plot]
table {%
12961 0.320149748915374
2463187 0.153047435493632
4971988 0.10405865957588
7555101 0.131047701391961
10038890 0.128145769833484
12643731 0.133565723487985
15186414 0.147631062478614
17699103 0.178163257258036
20312329 0.197865005449913
22817317 0.167218747905208
25378192 0.187762928306628
27897083 0.162969007289961
30544357 0.216200403233572
33024976 0.181677314777379
35551853 0.17474875554373
38087164 0.184539379192178
};
\addplot [thick, color0, forget plot]
table {%
1010 0.323997893417724
2448678 0.155463678950463
4961599 0.122579643794457
7547284 0.143412071185814
10033840 0.138256255867115
12610777 0.141977921182953
15181858 0.154805115428433
17705900 0.156658208072442
20357027 0.165734411304827
22803337 0.181993620682125
25365016 0.170933667774337
27897270 0.173364967368109
30542242 0.193244915834518
33007214 0.211173301781386
35528487 0.19921060678234
38056723 0.174783290219211
};
\addplot [thick, color1]
table {%
10918 0.385728141274114
2558265 0.466901378499138
5034129 0.722952716278309
7644862 0.917387894909792
10190417 0.957685403751605
12733012 0.939230929721486
15351765 0.937825786947
17772025 0.975780225763419
20447690 0.980067077911263
22901585 0.986126553530646
25457652 0.98791436233906
28070789 0.986009730233085
30517791 0.982029141801778
33111900 0.978665018322494
35623142 0.975557630110271
38157358 0.963596896089688
};
\addplot [thick, color1, forget plot]
table {%
2054 0.372663118923553
2533607 0.551022047045254
5001742 0.8344670642506
7603958 0.947415964772003
10145082 0.959844427277344
12692689 0.948778769584617
15318813 0.946865596554495
17743020 0.957260478626598
20435472 0.968948174606672
22896132 0.957233089389223
25456273 0.972080277674127
28068218 0.975315680407516
30513616 0.979559615404919
33134525 0.983941345503838
35646702 0.975960519578723
38179028 0.970332908509956
};
\addplot [thick, color1, forget plot]
table {%
2251 0.174745831901979
2539428 0.00410111560859664
4999591 0.0988514850570122
7555108 0.451220048854898
10075345 0.672891692982779
12614069 0.67867173299645
15257305 0.715511544485286
17630105 0.761382327537345
20288897 0.800861945658019
22851988 0.832789128777957
25383598 0.839452215818443
27980667 0.867901445037185
30488995 0.881595926453368
33046733 0.874663913490797
35576767 0.93020598394702
38250580 0.92363013462587
};
\addplot [thick, color1, forget plot]
table {%
13754 0.193065299530222
2561434 0.131280778034712
5037573 0.321366624624441
7600637 0.507370597410083
10121106 0.648023724405452
12654046 0.695734564252572
15294726 0.737074740005263
17653529 0.757226007755357
20300292 0.805634868867469
22859032 0.821892283781611
25388128 0.885155084759298
27981143 0.870453783357987
30489372 0.875003414924696
33037373 0.907473707439922
35585195 0.929293162292903
38255788 0.911050445503661
};
\end{axis}

\end{tikzpicture}
	\caption{\label{fig:attention-weights-2-2} Second sub-network.}
\end{subfigure}
\hfill
\begin{subfigure}{.32\textwidth}
\begin{tikzpicture}

\definecolor{color0}{rgb}{0.223529411764706,0.415686274509804,0.694117647058824}
\definecolor{color1}{rgb}{0.854901960784314,0.486274509803922,0.188235294117647}

\begin{axis}[
xlabel={Steps},
xmin=-1911728.9, xmax=40168526.9,
ymin=0, ymax=1,
width=\triplePlotWidth,
tick align=outside,
tick pos=left,
x grid style={white!69.01960784313725!black},
y grid style={white!69.01960784313725!black},
label style={font=\triplePlotAxesFontSize},
tick label style={font=\triplePlotTicksFontSize},
]
\addplot [thick, color0]
table {%
7481 0.343107029344096
2436497 0.48208290022431
4978934 0.475933645861318
7571552 0.52006778474709
10034867 0.582850454130558
12582158 0.639997547172536
15091545 0.590265715965118
17659633 0.591403717645492
20217858 0.630326449720547
22781692 0.563457789896715
25243546 0.595873537238199
27785792 0.602024180118483
30405588 0.688796755189848
32887349 0.576499948146368
35498132 0.619380619489786
38028752 0.646796535512413
};
\addplot [thick, color0, forget plot]
table {%
2008 0.352279171498135
2458066 0.475997380385496
4966947 0.34399200133001
7532012 0.48549504420071
10009771 0.537672591179308
12602705 0.594645776381397
15179732 0.58089269787976
17689136 0.606293816608611
20348512 0.62356443688123
22798967 0.603770654911948
25349559 0.587333345051969
27865724 0.64353810913033
30524536 0.690815465317834
32975315 0.650797440849171
35487287 0.664147374756408
38010775 0.628356775701647
};
\addplot [thick, color0, forget plot]
table {%
12961 0.341138826777236
2463187 0.510358439852493
4971988 0.773056234976257
7555101 0.809230602750876
10038890 0.836255524495637
12643731 0.843793510186552
15186414 0.835521869587176
17699103 0.805462803503481
20312329 0.786793419508019
22817317 0.820671384984799
25378192 0.799908559129696
27897083 0.824607217552686
30544357 0.776654385256046
33024976 0.812862135846206
35551853 0.821510975409036
38087164 0.812854619038226
};
\addplot [thick, color0, forget plot]
table {%
1010 0.339475646163478
2448678 0.505356912661081
4961599 0.730423608813622
7547284 0.794621925763408
10033840 0.822880976127855
12610777 0.833537857339842
15181858 0.824611396500561
17705900 0.827448877120258
20357027 0.82046018224774
22803337 0.804135096193566
25365016 0.816845324605402
27897270 0.813431722347183
30542242 0.800633278157977
33007214 0.782542797050089
35528487 0.796642945270348
38056723 0.822461142684473
};
\addplot [thick, color1]
table {%
10918 0.306526148861104
2558265 0.247128480161079
5034129 0.0860086446157611
7644862 0.0320580197930939
10190417 0.0135291821671403
12733012 0.0191582930697635
15351765 0.0204130759312197
17772025 0.0100066884505479
20447690 0.00798504694713055
22901585 0.00623827972543434
25457652 0.00668366789356853
28070789 0.00839787792659015
30517791 0.0145440269355673
33111900 0.0194218716010302
35623142 0.0232594354463873
38157358 0.0357146576385606
};
\addplot [thick, color1, forget plot]
table {%
2054 0.313559323248237
2533607 0.198088134328524
5001742 0.0363158232372519
7603958 0.0155991994107913
10145082 0.0112190805657794
12692689 0.015492639193932
15318813 0.0166224268914171
17743020 0.0182974806638679
20435472 0.0119750783825764
22896132 0.018946660200435
25456273 0.013804710765294
28068218 0.0127948061986404
30513616 0.0148941545648443
33134525 0.0135190241330426
35646702 0.0215596354124372
38179028 0.028950448236146
};
\addplot [thick, color1, forget plot]
table {%
2251 0.418972365181857
2539428 0.818983687896922
4999591 0.796452767921217
7555108 0.489865935812092
10075345 0.275240879694019
12614069 0.275264343754812
15257305 0.2480043542355
17630105 0.19738272449585
20288897 0.167535747730672
22851988 0.144951986468802
25383598 0.154137921039805
27980667 0.121763493461214
30488995 0.113305559889837
33046733 0.123201173043461
35576767 0.0690512098654198
38250580 0.0761654000409474
};
\addplot [thick, color1, forget plot]
table {%
13754 0.408550312723776
2561434 0.533762096154568
5037573 0.388318010803425
7600637 0.251630897172774
10121106 0.20693655793715
12654046 0.18185505361268
15294726 0.161036374141472
17653529 0.148704100061547
20300292 0.127147715501111
22859032 0.113543712731564
25388128 0.0814424421675878
27981143 0.0831871398094327
30489372 0.107402885407962
33037373 0.0834216570782692
35585195 0.0649674171675934
38255788 0.0871173403498653
};
\end{axis}

\end{tikzpicture}
	\caption{\label{fig:attention-weights-2-3} Third sub-network.}
\end{subfigure}
\caption{\label{fig:attention-weights} Smoothed average of the attention weights for $8$ different tasks of two domains. Grid World (blue) consists of grid world tasks $3$ to $6$ and Connect Four (orange) consists of the four connect-four/five tasks. The connect-four/five tasks are almost exclusively learned into the second sub-network, while the grid world tasks use mostly the third sub-network. We can see that there is a clear separation between the weights of the tasks of the two domains.}
\end{figure*}

To see whether the approaches can handle transfer between domains where the action spaces are not aligned, we take the second, third and fourth grid world task and switch their action dimensions, meaning that the agent goes to the left instead of the right, to the top instead of the bottom, and vice versa.
We combine these new tasks with the original grid world tasks $2, 3$ and $4$ and train the three different approaches to solve these six tasks simultaneously.
Figure~\ref{fig:swap-actions} compares the number of steps required to reach a score of $0.9$ on all tasks separately and on average. Our approach clearly outperforms PNN and Distral in the number of steps required the reach the target performance on all tasks. We see two explanations for this: either two of our sub-networks specialize to the two sets of tasks and allow fast transfer as such, or the task specific linear layer $(W_\tau,b_\tau)$ in our architecture effectively learns to invert the action space of some tasks such that two tasks from the two different sets look similar to a sub-network in our architecture. Most likely, the improvement is due to an entangled combination of both explanations.
The results of PNN are comparable to the previous experiment as PNN is also able to deal well with unaligned action spaces. In contrast to multi-task approaches however, PNN is bound to define a threshold for when to freeze the current task's network weights and move on to the next one. Further, a curriculum needs to be specified and tasks learned earlier cannot profit from knowledge discovered during learning later tasks. Therefore PNN ultimately learns slower in this setup than our approach.
Distral, an approach that aligns the action space between tasks, requires more steps for these six tasks than for randomly selected six tasks like in the previous experiment, as the distilled policy cannot deal with the three environments and their counterparts at the same time. This underlines our claim that, while other approaches are effective for multi-task learning in a controlled setup, our approach is able to deal with multiple tasks even if the action spaces are not aligned.

\subsection{Analyzing the Learned Attention Weights} \label{subsec:analyzeattweights}

To give an insight in how tasks are separated into sub-networks we take a sub-set of the grid world tasks where we expect negative transfer when knowledge is shared. More specifically, we take tasks $3-6$ and $7-10$. Note that these two sets of tasks are visibly indistinguishable and equivalent apart from the fact that bonus objects yield negative rewards and penalty objects positive rewards in the second set.
In Figure~\ref{fig:attention-weights-grid-world} we plot a smoothed average of the attention weights $w_i(s,\tau)$ of each task $\tau$ for all sub-networks $i\in  \{1,2,3\}$. As can be seen in the figure, our architecture discovers that three sub-networks are not needed for these six tasks and learns to discard one of them. Further, one can see a tendency that one set of tasks is learned into one of the sub-networks while the other set of tasks is learned into the other remaining sub-network. Note however, that this distinction is not sharp since there is still a lot of transfer possible between the two sets of tasks, i.e., the agent has to stay on the board and find the target in both sets.
This brings us to the interesting question how the distribution of the weights would look like if one uses two sets of tasks from completely unrelated domains. To answer this question, we train our model on connect-four/five and on grid world tasks.
Figure~\ref{fig:attention-weights} shows the weighting of the sub-networks when trained on those tasks. Clearly, the second sub-network learns to specialize on the connect-four/five task.
Further, even though the connect-four/five and grid world tasks are unrelated to each other, parts of the \quotes{connect-four-knowledge} is used for the grid worlds while the non-overlapping state-action correlations are safely learned in a separate sub-network.
Again, one of the sub-networks is left almost unused by all tasks, i.e., the model automatically learned that there are more sub-networks than needed for the two task domains.

\section{Conclusion}

We present a multi-task deep reinforcement learning algorithm based on the intuition of human attention. We show that knowledge transfer can be achieved by a simple attention architecture that does not require any a-priori knowledge of the relationship between the tasks. We show that our approach achieves transfer comparable to state of the art approaches as the number of tasks increases while using substantially fewer network parameters.
Further, our approach clearly outperforms Distral and PNN when the action space between tasks is not aligned, since the task-specific weights and specialized sub-networks can account for this discrepancy.
In future work, we plan to apply our approach to more complex tasks by incorporating recent, more resource efficient algorithms like \cite{IMPALA,PopArt-IMPALA}.


\bibliographystyle{splncs04}
\bibliography{biblio}

\end{document}